\documentclass[dvipsnames]{article}
\usepackage{amsmath,graphicx,mlspconf}
\usepackage{booktabs} 

\copyrightnotice{979-8-3503-2411-2/25/\$31.00 {\copyright}2025 IEEE}

\toappear{2025 IEEE International Workshop on Machine Learning for Signal Processing, Aug.\ 31-- Sep.\ 3, 2025, Istanbul, Turkey}

\title{\textbf{GEMS}: \underline{G}roup \underline{E}motion profiling through \underline{M}ultimodal \underline{S}ituational understanding}
\name{ %
   Anubhav Kataria$^{1}$%
   \qquad Surbhi Madan$^{2}$%
   \qquad Shreya Ghosh$^{3}$
   \qquad Tom Gedeon$^{3}$%
   \qquad Abhinav Dhall$^{4}$\thanks{This work was supported by resources provided by the Pawsey Supercomputing Research Centre’s Setonix Supercomputer (https://doi.org/10.48569/18sb-8s43), with funding from the Australian Government and the Government of Western Australia.
}%
}
\address{%
   $^{1}$ Kroop AI, $^{2}$ IIT Ropar, $^{3}$ Curtin University, $^{4}$ Monash University\\ 
   \small Emails: \texttt{anubhav.kataria@kroop.ai}, \texttt{surbhi.19csz0011@iitrpr.ac.in}, \\
   \small \{\texttt{shreya.ghosh}, \texttt{tom.gedeon}\}\texttt{@curtin.edu.au}, \texttt{abhinav.dhall@monash.edu}\\%
   }

\usepackage{graphicx}
\usepackage{amsmath}
\usepackage{amssymb}
\usepackage{booktabs}
\usepackage{url}
\usepackage{multirow}
\usepackage{multicol}
\usepackage{algpseudocode}
\usepackage{listings}
\usepackage{tablefootnote}
\usepackage[dvipsnames]{xcolor}
\usepackage{enumitem}
\usepackage{algorithm}
\usepackage{algpseudocode}
\usepackage{sidecap,wasysym,array,tabularx,caption}

\algrenewcommand{\Return}{\State\algorithmicreturn~}
\usepackage{colortbl}
\usepackage[normalem]{ulem}
\usepackage{relsize,amsfonts}
\usepackage{import}
\definecolor{mygray}{gray}{.9}
\usepackage{tcolorbox}
\newtcolorbox[list inside=prompt,auto counter,number within=section]{prompt}[1][]{
    colbacktitle=black!60,
    coltitle=white,
    fontupper=\footnotesize,
    boxsep=5pt,
    left=0pt,
    right=0pt,
    top=0pt,
    bottom=0pt,
    boxrule=1pt,
    title={#1},
    #1, 
}
\begin{document}

\maketitle

\begin{abstract}

Understanding individual, group and event level emotions along with contextual information is crucial for analyzing a multi-person social situation. To achieve this, we frame emotion comprehension as the task of predicting fine-grained individual emotion to coarse grained group and event level emotion. We introduce \textbf{GEMS} that leverages a multimodal swin-transformer and S3Attention based architecture, which processes an input scene, group members, and context information to generate joint predictions. Existing multi-person emotion related benchmarks mainly focus on atomic interactions primarily based on emotion perception over time and group level. To this end, we extend and propose \textbf{VGAF-GEMS} to provide more fine grained and holistic analysis on top of existing group level annotation of VGAF dataset~\cite{sharma2019automatic}. GEMS aims to predict basic discrete and continuous emotions (including valence and arousal) as well as individual, group and event level perceived emotions. Our benchmarking effort links individual, group and situational emotional responses holistically. The quantitative and qualitative comparisons with adapted state-of-the-art models demonstrate the effectiveness of GEMS framework on VGAF-GEMS benchmarking. We believe that it will pave the way of further research. The code and data is available
at: \url{https://github.com/katariaak579/GEMS}
\end{abstract}

\newcommand{\cem}[1]{\textcolor{blue}{cem: #1}}

\section{Introduction}
\label{sec:introduction}

Comprehensive situational interpretation requires emotional understanding over time as human emotion is associated with multiple entities (e.g. persons, emotion, actions, interaction and scenes) interacting dynamically over time. Although progress has been made in several benchmarking efforts for individual level~\cite{yang2024emollm}, context level~\cite{venkatraman2024multimodal}, group level~\cite{ghosh2018automatic}, event level~\cite{malhotra2023social} emotion understanding, however, current state-of-the-art models yet struggle to discriminate between multimodal as well as multiple entities (e.g. persons, actions, and scenes) based emotion understanding such as in a birthday situation where the situation level emotion is happy/excited. However, each individual may not be expressing happy expression throughout the whole situation. Group-level emotional understanding has a wide range of potential applications, including providing automatic feedback for lectures~\cite{whitehill2014faces}, monitoring student engagement in classrooms~\cite{kaur2018prediction}, event detection~\cite{vandal2015event}, group cohesion~\cite{ghosh2019predicting,ghosh2020automatic}, surveillance, ranking images from events~\cite{dhall2015automatic}, and summarizing events~\cite{dhall2015automatic}. 

\begin{figure}
    \centering
    \includegraphics[width=\linewidth]{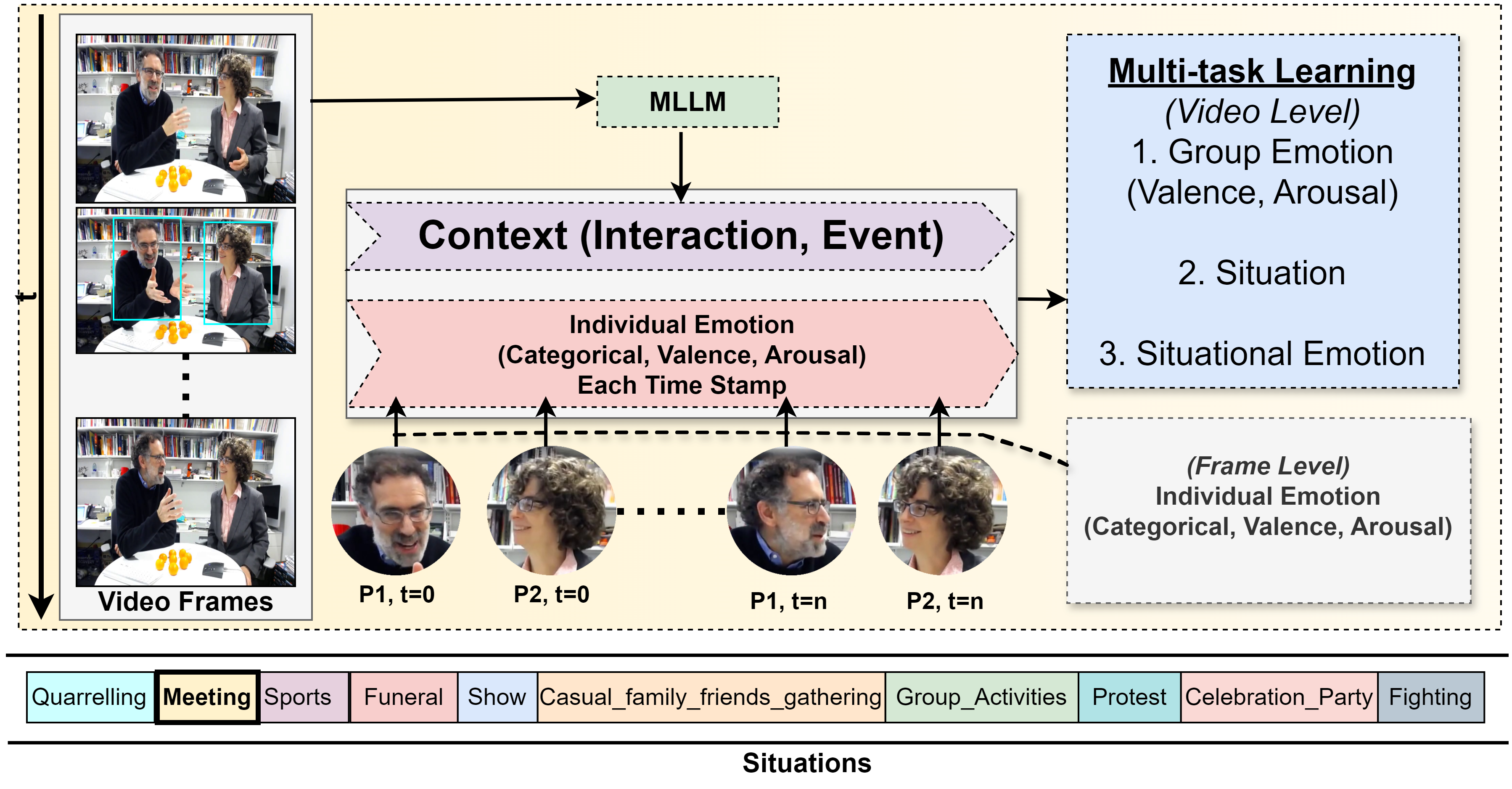}
    \vspace{-5mm}
    \caption{\small \textbf{GEMS.} A brief overview of GEMS framework. Given any of the defined situations (eg: Celebration-party, Fighting), the framework aims to link participant's individual emotion, group emotion, situational emotion.}    
    \label{fig:teaser}
    \vspace{-8mm}
\end{figure}

Previous work on group-level emotion analysis~\cite{ghosh2018automatic, mou2016alone, huang2024survey} primarily focused on individual-level analysis followed by aggregation, but lacked annotations capturing temporal dynamics and situational context critical for accurate group-level perception. To this end, we propose VGAF-GEMS, a new benchmark to objectively test different levels of emotion perception and evaluate the compositional ability of individual to group and event level emotion understanding through the lens of time over a situation (See Fig.~\ref{fig:teaser}). We adopt VGAF~\cite{sharma2019automatic}, a group level emotion recognition dataset that consists of coarse grained group level emotion annotations with event level tags. Sourced from youtube clips, VGAF videos present challenging scenarios with frequent shot changes, fast action sequences, multi-event complex situations, role switching between the speaker and listener, all entangled through time. We parse these complex videos with dense individual level emotion annotations capture multiple and changing aspects of the situations in the video along with group level pooling with situational information. We build a rich annotations to create a challenging and high-quality benchmark test suite that requires binding the correct emotion concepts beyond a individual or group level emotion.

\begin{itemize} [topsep=1pt,itemsep=0pt,partopsep=1ex,parsep=1ex,leftmargin=*]
    \item We propose \textbf{G}roup \textbf{E}motion profiling through \textbf{M}ultimodal \textbf{S}ituational understanding aka \textbf{GEMS} that covers depth and breadth of emotional profiling in a given situation. 
    
    \item To the best of our knowledge, we are the first to propose \textbf{VGAF-GEMS}, a densely annotated group emotion profiling benchmark that covers the compositionality of individual, situation, contextual aspects of people interacting in a short video. Please note that we annotate images from a third-person perspective following the literature and we incorporate a novel semi-automatic interface-based data annotation strategy using Emolysis~\cite{ghosh2023emolysis} toolkit.

    \item We perform comprehensive analysis and benchmark the proposed dataset utilizing state-of-the-art emotion recognition algorithms. We benchmark the datasets for two different learning paradigms, \textit{i.e.}, zero-shot and fully supervised. The significant drop in performance (i.e., $\sim \textbf{54.80}$ w.r.t VGAF dataset with supervised framework) directly indicate that our dataset will be a valuable asset to pursue further research in this domain.
\end{itemize}

\section{Background}
\label{sec:background}

Prior work in group-level emotion analysis can be divided into three main categories: \textit{bottom-up}, \textit{top-down}, and \textit{hybrid} approaches. \textit{Bottom-up approaches} focus on analyzing individual group members' emotions and aggregating their contributions to assess the overall group mood~\cite{MoodMeter}. In contrast, \textit{top-down approaches} begin with global factors affecting the group and examine how these factors shape the perception of the group's emotional state~\cite{barsade1998group}. \textit{Hybrid approaches} combine both group-level and individual-level information to estimate the group's emotional state~\cite{mou2016alone,ghosh2018automatic,sharma2021audio,sharma2019automatic}. A comprehensive description of the prior works is indicated at~\cite{huang2024survey}. 

\begin{figure*}[!h]
    \centering
    \includegraphics[width=0.33\linewidth,height=35mm]{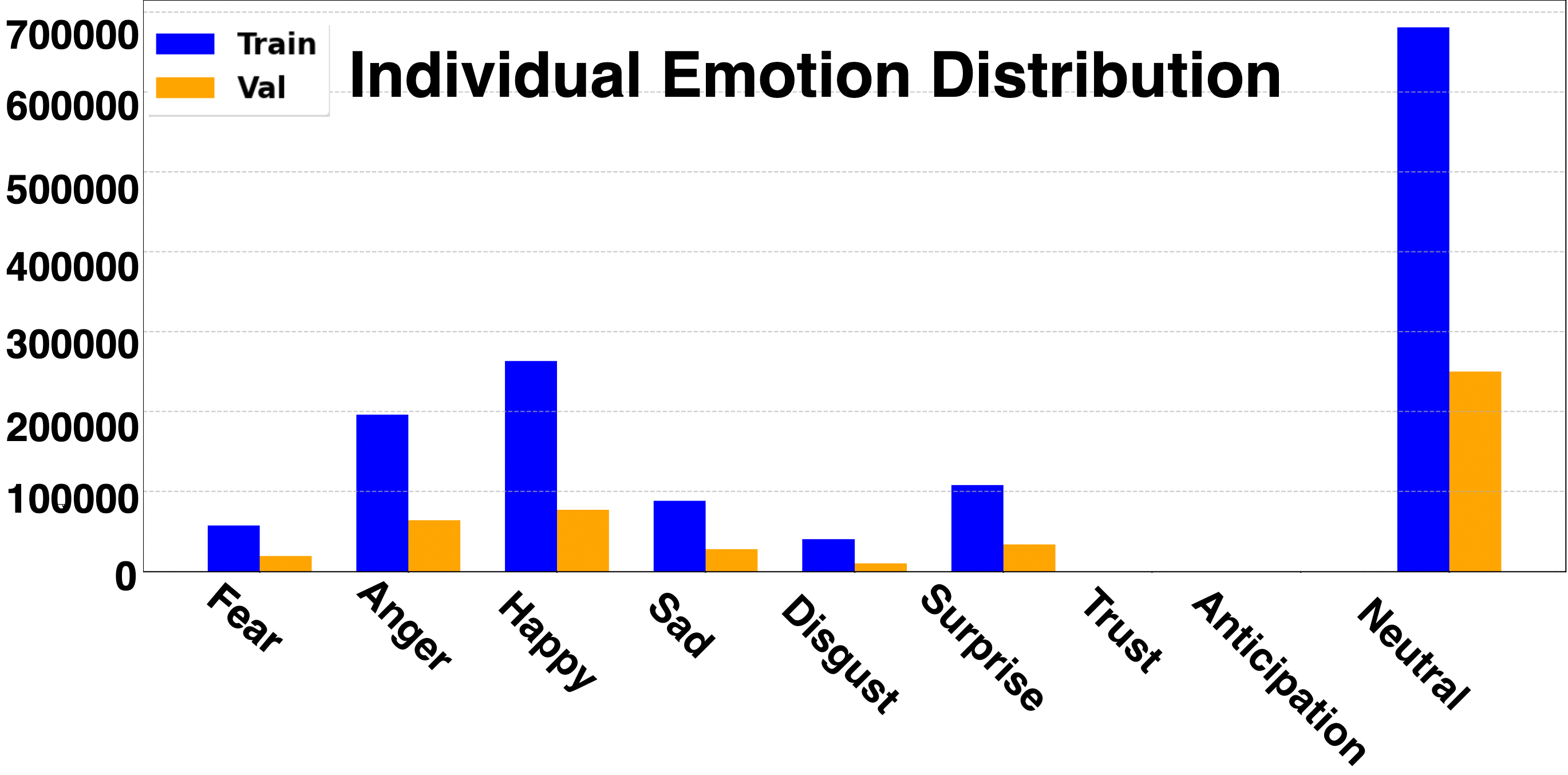}
    \includegraphics[width=0.33\linewidth,height=35mm]{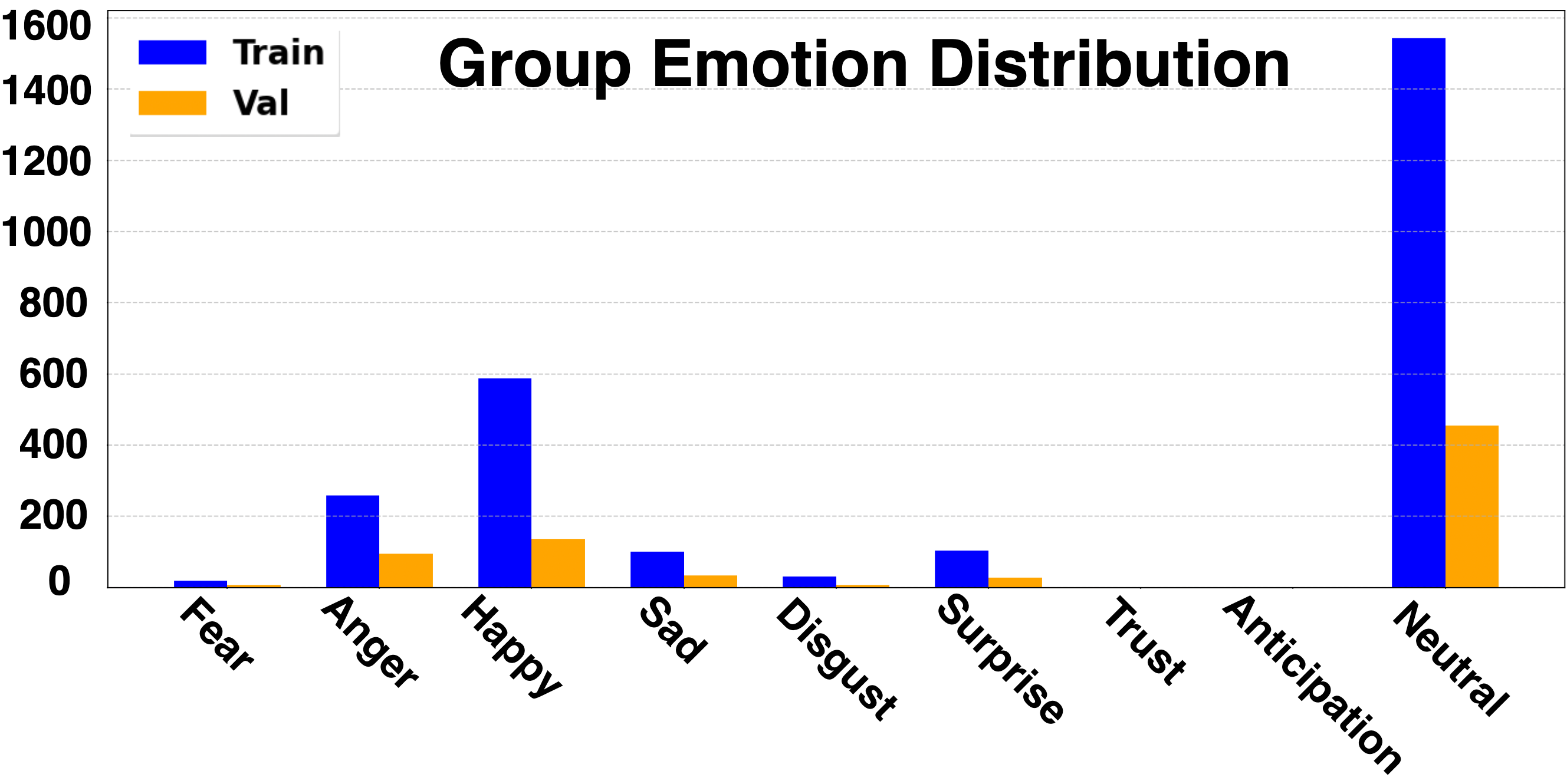}
    \includegraphics[width=0.33\linewidth,height=35mm]{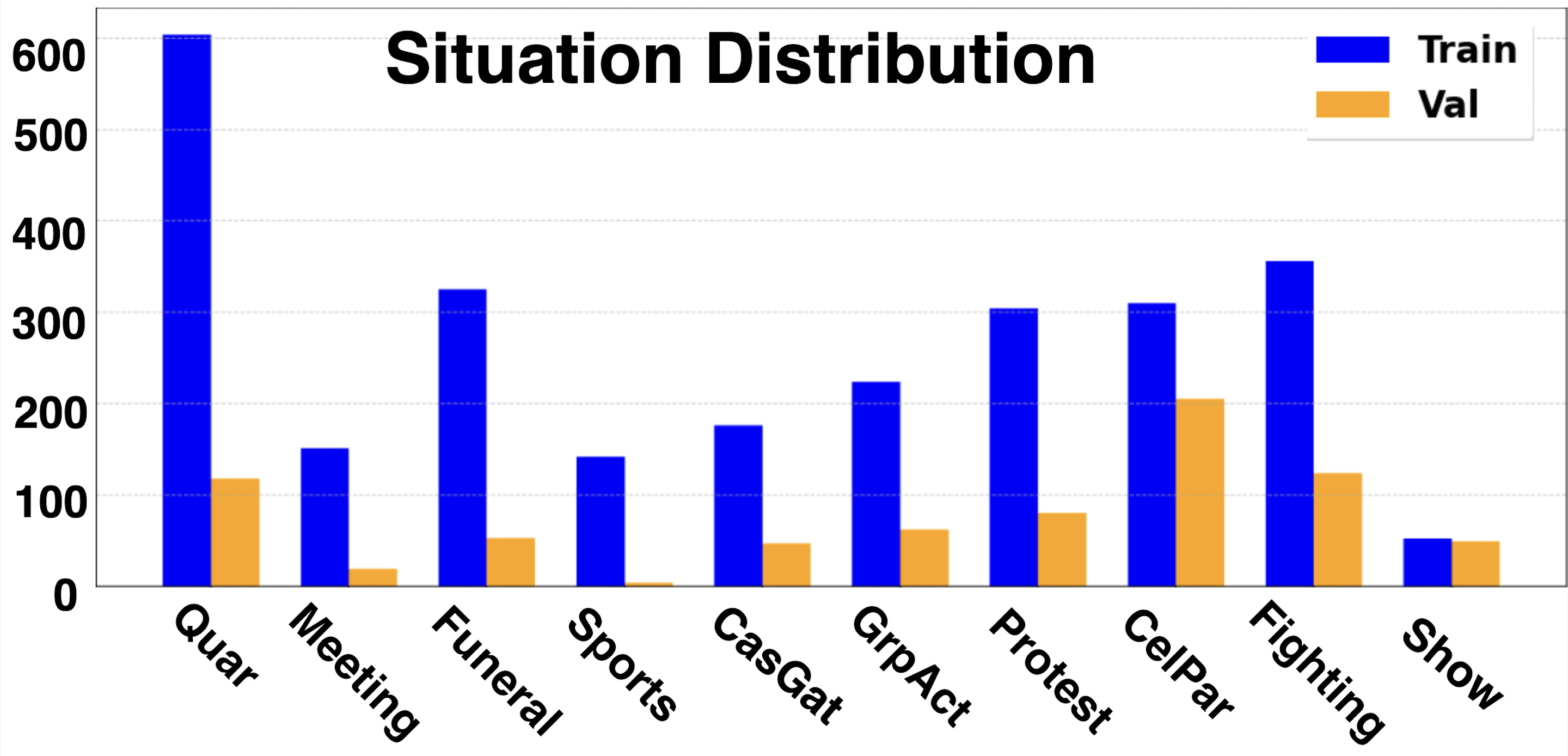}
    \vspace{-4mm}
    \caption{\small \textbf{Data statistics for the VGAF-GEMS dataset.} The figure presents the class-wise distribution of individual emotions, group emotions, and situational contexts in the proposed dataset. Note, Quar: Quarrelling, CasGat: Casual Gathering, GrpAct: Group Activities and CelPar:Celebration Party. The y-axis indicates the number of samples corresponding to each class.}
    \label{fig:data}
    \vspace{-6mm}
\end{figure*}

An initial work~\cite{rel_p3_7} was proposed back in 2015 on automatic recognition of group emotions using web crawled static images. These images have groups of individuals and were labeled according to six levels of happiness intensity (i.e in arousal axis of emotion). Later works in this domain expanded the research into a multimodal analysis using audio visual inputs along with extending it to valence axis of emotion: i.e. negative, neutral, and positive~\cite{rel_p3_9,rel_p3_16,rel_p3_15,rel_p3_13,rel_p3_47}. Related research has examined multi-level emotion analysis from different perspectives. Yilmaz et al. [22] investigated sentiment analysis with multiple emotion labels across 100 languages. Their work shows how multiple emotions can exist simultaneously in a single piece of content - much like how our videos contain multiple individuals each expressing different emotions that together create a complex group emotional state. In another study, Giritlioglu et al.~\cite{giritliouglu2021multimodal} analyzed video personality traits by looking at both what people do and how they look. This is similar to our work, where we combine individual emotions to understand group emotions using multiple types of information. However, these efforts faced two significant limitations: (1) the works did not consider the temporal dynamics and shifts in group emotions across consecutive video segments. As a result,  the emotion annotations were temporally independent, and (2) the papers did not contextualize group emotions within the framework of group interactions. To address these issues, Lei et al.~\cite{rel_p3_6} annotated 2-minute segments of group interactions, incorporating temporal context from interaction videos. Nonetheless, the extended duration of these segments hindered a more nuanced analysis of the dynamics of group emotions. Another study~\cite{rel_p3_29}  was conducted in a simpler context, where participants merely watched videos and displayed emotions through facial and bodily gestures, rather than in genuine interaction settings. More recently, Wang et al.~\cite{rel_p3_14} introduced a graph neural network architecture to model group emotions using static images. However, this work was constrained by the absence of annotations on the dynamics of group emotions, which impose different constraints over group level inference.

\section{VGAF-GEMS}
\textit{VGAF-GEMS} is a large-scale, video-based group emotion recognition dataset, developed from the existing VGAF dataset~\cite{sharma2019automatic}. It consists of 4,183 video samples, divided into 2,661 training, 766 validation, and 756 testing samples. The videos, each 5 seconds long with frame rates ranging from 13 to 30 frames per second, with a very coarse ground truth label in valence axis of emotion (positive, negative and neutral). To this end, the extended version of VGAF-GEMS provides continuous as well as categorical emotion for each individual per frame, along with event-based labels marking key moments and associated emotion. Additionally, we use MLLM models such as VideoGPT~\cite{maaz2023video} to generate descriptive annotations for each video, providing contextual information about the scenes. Group-level emotion labels (positive, negative, or neutral) are also included to capture collective emotional states. These enhancements improve the dataset's utility for both individual, situational and group-level emotion analysis in multi-person video data.

\noindent \underline{\textbf{Data Annotation Pipeline}}
We annotate the VGAF~\cite{sharma2019automatic} using Emolysis toolkit~\cite{ghosh2023emolysis} in a semi-automatic way mentioned below. First, before starting the annotation process, we compare the annotation agreement of 100 videos as compared to the Emolysis framework based automatic annotation. Among 100 videos, mainly the annotation of individual level emotion, valence and arousal need to be corrected in 67.2\% of the frames.

\noindent \textbf{Individual Emotion.}
Emolysis toolkit uses off-the-shelf HSEmotion~\cite{savchenko2022hsemotion} model trained on Affectnet dataset to predict categorical emotion, continuous arousal and valence scores. We initialize the individual level; emotion values using Emolysis toolkit. Further to ensure accuracy, we incorporated a human-in-the-loop approach, where human evaluators cross-verified the emotional scores for each individual. Two expert annotators parse through the annotation process to ensure the individual level emotion annotation. There is a third annotator who only parse through the tie cases.

\noindent \textbf{Group Level Emotion.} Similar to individual level annotation, we incorporate label initialization using Emolysis toolkit followed by human-in-the-loop label refinement.


\noindent \textbf{Situational Emotion.}
For situation-based emotion labeling, we frame this as a multi-label classification task. Following literature~\cite{malhotra2023social}, each video clip is categorized into one of ten distinct situations: \emph{Quarrelling}, \emph{Meeting}, \emph{Funeral}, \emph{Sports}, \emph{Casual-gathering}, \emph{Group-activities}, \emph{Protest}, \emph{Celebration-party}, \emph{Fighting}, and \emph{Show}. Based on social sciences~\cite{nezlek2008emotions}, we use a corresponding set of ten situational emotions: Active, Enthusiastic, Anxious, Angry, Attentive, Distressed, Sad, Relaxed, Rejected, and Happy. To create the ground truth for training, each situation is mapped to its predefined set of expected emotions, which is then encoded as a 10-dimensional multi-hot vector. For a video labeled with a specific situation, the positions in the vector for its mapped emotions were set to 1, while all others were set to 0. For instance, a video categorized as 'Quarrelling' receives a target vector where the indices for 'Active', 'Enthusiastic', 'Anxious', and 'Angry' are set to 1. The situation labels were manually annotated, and a label space survey with 10 participants showed an average perception agreement of 69\% for the situation and 67\% for the situational emotion, respectively.

\begin{figure}[t]
\centering
\includegraphics[width = 0.95\linewidth]{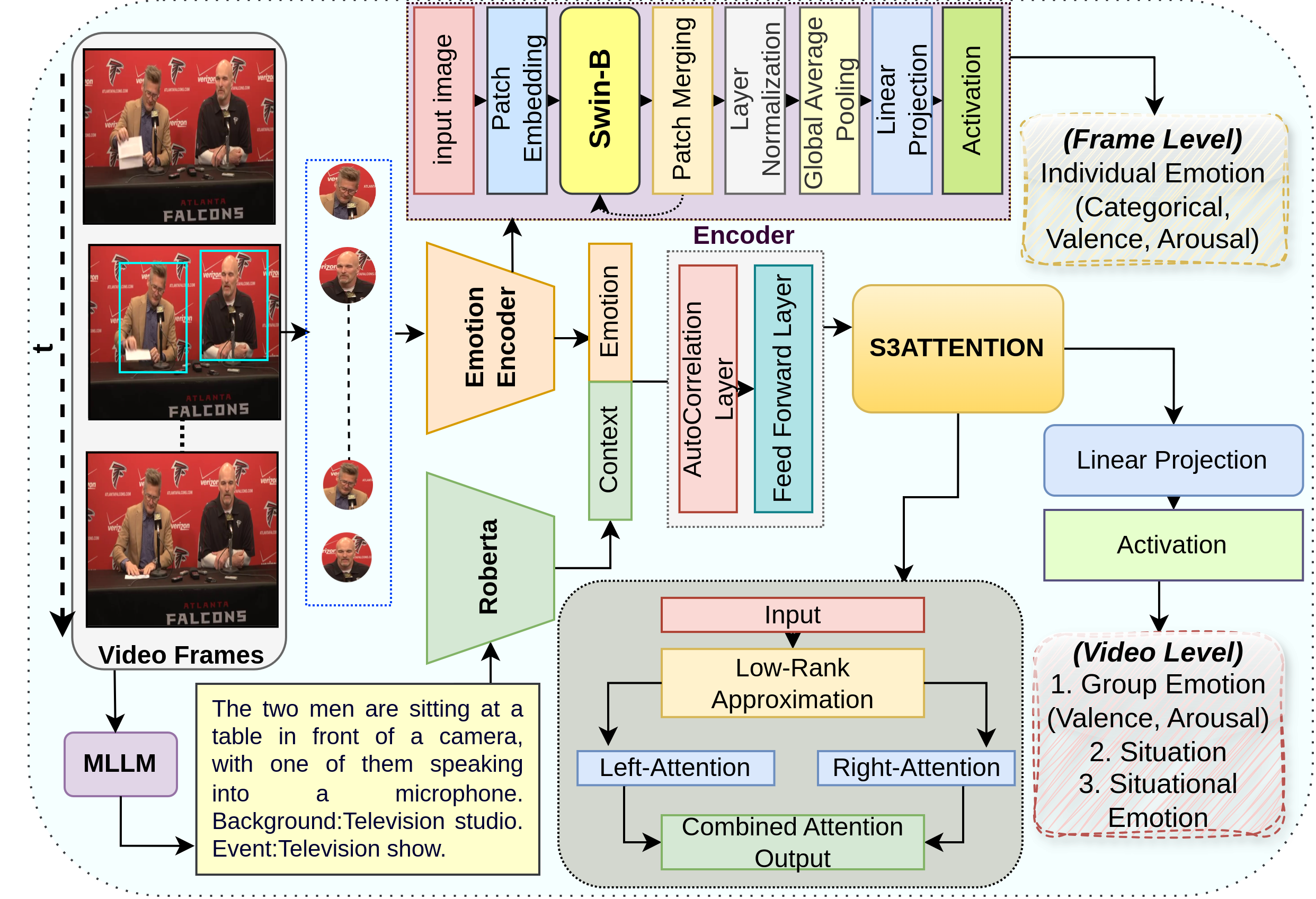}
\caption{\textbf{GEMS Overview.} Given a video input, GEMS parse individual level emotion embedding via emotion encoder, and contextual embedding via MLLM. contextual and individual level information is fused by S3 Attention mechanism to get group level and situation level information.} 
\label{fig:pipeline}
\vspace{-7mm}
\end{figure}

\begin{table*}[t]
\caption{\textbf{VGAF-GEMS Benchmarks.} We compare the performance of off-the-shelf models on our proposed VGAF-GEMS dataset. \textit{Please note that VGAF test set labels are not publicly available. Thus, we utilize only train and validation partition.}}
\label{tab:results}
\scalebox{0.65}{
\begin{tabular}{l|l||l|ccccc|ccccc|c|c}
\toprule[0.4mm]
\rowcolor{mygray}\textbf{Data} &\textbf{Supervision} &  \textbf{Method} & \multicolumn{5}{c|}{\textbf{Individual-Emotion}} & \multicolumn{5}{c|}{\textbf{Group-Emotion}} & \multicolumn{1}{c|}{\textbf{Situational-Emotion}} & \multicolumn{1}{c}{\textbf{Situation}}\\
\rowcolor{mygray}\textbf{} & \textbf{} & \textbf{} & \textbf{9-Class} & \multicolumn{2}{c}{\textbf{Valence}} & \multicolumn{2}{c|}{\textbf{Arousal}}& \textbf{9-Class} & \multicolumn{2}{c}{\textbf{Valence}} & \multicolumn{2}{c|}{\textbf{Arousal}}& \textbf{10-class} & \textbf{10-class}\\
\rowcolor{mygray}\textbf{} & \textbf{} & \textbf{} & \textbf{Acc.(\%) $\uparrow$} & \textbf{MSE$\downarrow$}& \textbf{CCC$\uparrow$} & \textbf{MSE$\downarrow$} & \textbf{CCC$\uparrow$} & \textbf{Acc.(\%) $\uparrow$} & \textbf{MSE$\downarrow$}& \textbf{CCC$\uparrow$} & \textbf{MSE$\downarrow$}& \textbf{CCC$\uparrow$} & \textbf{Acc.(\%) $\uparrow$}& \textbf{Acc.(\%) $\uparrow$} \\
\hline\hline
 \multirow{4}{*}{\rotatebox[origin=c]{90}{\begin{tabular}{c}\textbf{VGAF-}\\\textbf{GEMS}\end{tabular}}} &  \multirow{1}{*}{Zero-shot}& Video-LLM~\cite{maaz2023video} & - & - & - & - & - & 01.17 & 0.0105 & 0 & 0.0075 & 0 & 15.77 & 12.70 \\ \cline{3-15}
&   & MiniCPM-V~\cite{yao2024minicpm} & - & -& -& - & - &33.86 &0.0372 &0.0862 &0.0861 & 0.0408&52.16 &13.01 \\ \cline{2-15}
 &\multirow{5}{*}{Supervised} & Visual  & 26.62 & 0.0392 & 0.1067 & 0.0487 & 0.0811 & 09.75 & 0.0098 &  0.5121 & 0.0285 & 0.0045 & - & - \\ \cline{3-15}
 &   & w/o S3  & -  & - & - & - & -& 09.46 & 0.0558 & 0.0267 & 0.0489 & 0.0857 & 01.07 & 06.53 \\ \cline{3-15}
 &   &  GEMS (Ours)  & 26.62 & 0.0392 & 0.1067 & 0.0487 & 0.0811  &55.32 & 0.0174  & 0.095 & 0.0104 & 0.764 &  55.05 & 15.50 \\ 
\bottomrule[0.4mm]
\end{tabular}}
 \vspace{-6mm}
\end{table*}

\noindent \underline{\textbf{Dataset Statistics}}
\noindent \textbf{Class-wise Distribution of label space} We split the dataset into two nonoverlapping sets: train and validation (similar to Original VGAF partition). In each set, we have the following statistics: 2,661 training, 766 validation videos (5 sec duration). The overview of classwise relabelled data distribution is presented in Figure~\ref{fig:data}. 

\noindent \textbf{Inter-annotator Agreement.} We have computed agreement among annotators using Cohen’s Kappa ($\kappa$) measure~\cite{mchugh2012interrater}. To this end, we computed $\kappa$ between individual emotion, valence, arousal, group emotion, situation and situational emotion labels are 0.63, 0.70, 0.65, 0.8, 0.76, 0.72.

\begin{prompt}[title={Prompt \thetcbcounter: Video-ChatGPT Interaction}]
\textbf{System:} Initialize Video-ChatGPT.

\textbf{Human:} \{$<$Interaction$>$ What are the interactions between the people in the video? Your answer should be one or multiple of the following: {interactions categories}. Please think and list all possible answers.\}

\textbf{AI:} \{EXAMPLE OUTPUT is the interaction class e.g. insults, thanks, helps, disagrees with, walks with, jokes, gives, compliments etc.\}

\textbf{Human:} \{$<$Scene$>$ What is the background of the groups of people in the video? Your answer should be one of the following: {Scenes categories}. Please think and generate only one word as the answer.\}

\textbf{AI:} \{EXAMPLE OUTPUT is scene class.\} 

\textbf{Human:} \{$<$Event$>$ What are the event of the group of people in the video? Your answer should be one or multiple of the following: {event categories}\}

\textbf{AI:} \{EXAMPLE OUTPUT is event class.\} 

\textbf{Human:} \{$<$Relationship$>$ What are the relationship  of the group of people in the video? Your answer should be one or multiple of the following: {relationship  categories}\}

\textbf{AI:} \{EXAMPLE OUTPUT is relationship class.\}
\end{prompt}

\section{Benchmark and Metrics}
The VGAF-GEMS benchmark framework, as depicted in Figure~\ref{fig:pipeline}, is designed to establish a connection between individual emotions, scene interactions, and event understanding, enabling the prediction of multiple labels from group videos. 

\noindent \underline{\textbf{Baseline Details.}} Given an input video, there are two input branches; one for processing visual input in terms of frames and another is vision-language input processing via LLM. 

\noindent \textbf{1. Visual Encoder.} Each frames of a given input video is processed by MTCNN face detector~\cite{zhang2016joint} to identify all individual faces, which are then passed through an \textit{Emotion Encoder} to generate emotional embeddings for each person. The backbone of \textit{Emotion Encoder} is SWIN-B~\cite{liu2021swin} model, which takes the detected faces as input and classifies it into 9-basic emotion categories~\cite{savchenko2022hsemotion}. The Emotion Encoder consists of multiple SWIN transformer blocks, patch merging, global average pooling, and linear layer with softmax activation containing 9 output neurons. The model is pre-trained on VGAF-GEMS individual emotion benchmarks. After training, we freeze this part and extract individual level emotion embedding from the penultimate layer of the SWIN-B network.

\noindent \textbf{2. MLLM Encoder.} The input video is also fed into a Multimodal Large Language Model (MLLM) using prompts to infer context information in terms of events, background, and interactions. We use off-the-shelf Video-ChatGPT~\cite{maaz2023video} to generate contextual descriptions in terms of  interaction, scene, event and relationship for each video to form a cohesive narrative. The output of MLLM produces contextual descriptions, which is further encoded into vector format by RoBERTa encoder~\cite{liu1907roberta}. RoBERTa builds on top of BERT with modified hyperparameters space and trained with larger mini-batches and learning rates to capture context information. The prompts are described below:

To compare MLLM-based inference with human perception, we conduct a user study with 5 participants (excluding the authors) on 20 videos, which shows high agreement ($\sim$85\%). However, we acknowledge that MLLM may be vulnerable to hallucination.


\noindent \textbf{3. Fusion and S3Attention.} The individual emotional and video level contextual embeddings are concatenated and passed through an encoder block containing auto-correlation layers, capturing spatio-temporal dependencies, and a feed-forward layer. The final embeddings are processed through the S3Attention module. The S3Attention module~\cite{zhou2022fedformer} address the challenges in traditional attention mechanisms for long sequences. It utilizes a smoothing block that combines global and local information to reduce noise and improve token clarity. Additionally, matrix sketching selects key data by focusing on important rows and columns. The Low-rank approximation method reduces computational complexity by focusing attention on short sub-sequences instead of entire long sequences. As shown in the figure~\ref{fig:pipeline}, left and right attention independently compute row and column attention, respectively, before the final output is used for multi-level predictions.

\noindent \textbf{4. Multitask Learning and Label Space.} Finally, the whole module is trained as multitask learning way where the output space at group level are group emotion (continuous and discrete), situational emotion (discrete) and situation.

\begin{figure*}[t]
    \centering
    \includegraphics[width=0.95\linewidth]{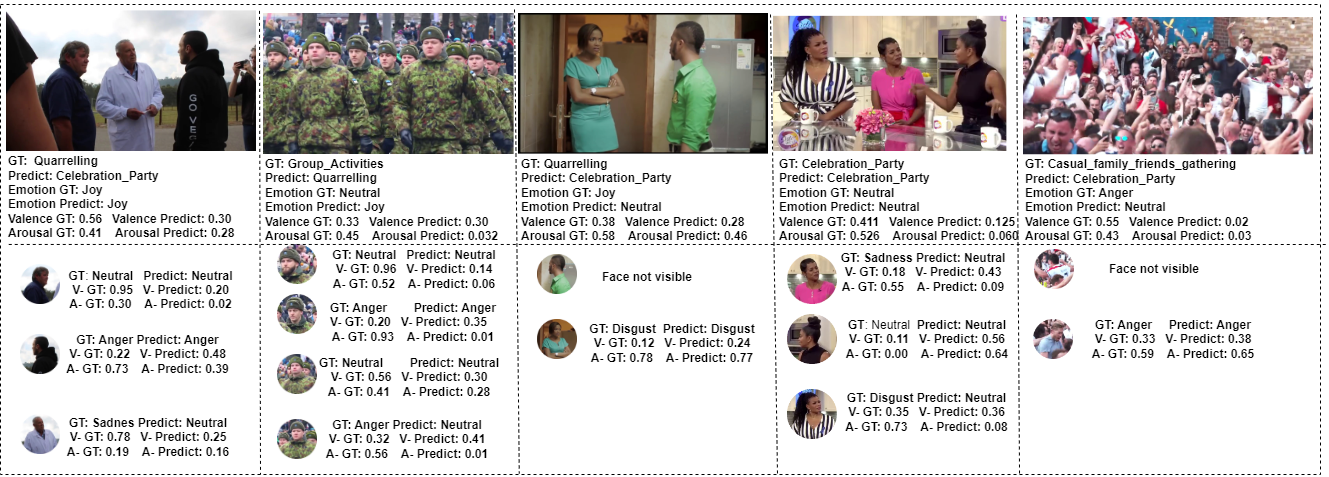}
    \caption{Qualitative analysis of the GEMS framework. Here, GT: Ground Truth and Predicted: Predicted from GEMS.}
    \label{fig:qualitative}
    \vspace{-6mm}
\end{figure*}

\noindent \underline{\textbf{Experimental Setup}}
\noindent \textbf{Implementation Details.} We conduct our experiments on an Nvidia A100 GPU with 40 GB of memory as well as Pawsey server. The network is trained using the Adam optimizer with a learning rate of 0.0001. We set the number of epochs to 1000 and apply early stopping with a patience parameter of 30 and batch size 32. we use mean absolute error and mean squared error as the loss functions for continuous emotion predictions (valence and arousal). For classification tasks such as event classification, group emotion classification, and others, categorical cross-entropy loss is applied with different neuron counts based on the number of classes. 

\noindent \textbf{Evaluation Metrics.} Following the standard evaluation protocol~\cite{dhall2017individual,sharma2019automatic,sharma2021audio,ghosh2018automatic}, we utilize class-wise accuracy (for discrete emotion), CCC and MSE (for continuous emotion) as evaluation metrics. \textit{Please note that VGAF test set labels are not publicly available. Thus, we utilize only train and validation partition.}

\noindent \textbf{Experimental Settings.} We evaluate the benchmark with two different setups: \textit{Zero-shot.} Here, we evaluate the zero shot capability of Large language models. \textit{Supervised.} Here, we train the model using supervised learning paradigm.



\section{Results and Discussion} 
\noindent \textbf{Quantitative Analysis.} The results of proposed VGAF-GEMS benchmark is presented in Table~\ref{tab:results}. 
The zero shot capability of the state-of-the-art video-LLM is quite weak. For group level inference, it performs around 1.17\% for 9-class group emotion, 15.77\% for situational emotion understanding. 
In supervised benchmark as well, the model performs notably low in multi task learning.
Please note that the addition of S3 attention module boosts the performance quite a lot ($9.46\% \rightarrow 54.80\%$). Please note that we are the first to propose the emotional compositionality over video in terms of individual, group and situational emotion. Thus, publicly available benchmarks are not comparable.  

\noindent \textbf{Qualitative Analysis.} The performance comparison of GEMS model on VGAF-GEMS benchmark is shown in Figure~\ref{fig:qualitative}. The results indicate that our dataset is more challenging and requires a more robust algorithm that aligns individual, group and situation level emotion. 

\noindent \textbf{Discussion.} In spite of having a rich literature in group level emotion understanding~\cite{dhall2015automatic,ghosh2018automatic,ghosh2018role,sharma2019automatic}, Emotional compositionality through the lens of time is still an unexplored topic. Also, group level emotion highly dependent on context rather than facial contribution. Here, we explore this topic from a holistic perspective to encode how emotion perception changes from single person to group and situation.

\section{Conclusion} 
This paper introduces VGAF-GEMS, a densely labeled dataset designed for analyzing emotional compositionality. Benchmarking the dataset using state-of-the-art methods reveals a notable drop in performance, highlighting the dataset's potential to play a key role in comprehensive emotion profiling for algorithm development. \textbf{Ethical Impact Statement.} We respect the licenses of the deployed components used in Emolysis~\cite{ghosh2023emolysis} toolkit and VGAF~\cite{sharma2019automatic} dataset. The use of an existing face detection library~\cite{Vladmandic,zhang2016joint} may introduce potential bias in the model. We plan to address these limitations in future updates. \textbf{Broader Impact.} We believe VGAF-GEMS can serve as a valuable benchmark for the multimedia research community, aiding in the development of algorithms focused on human-human interaction in real-world settings. Its rich and explainable data offers valuable contextual information for various real-world applications.


\small
\bibliographystyle{IEEEbib}
\bibliography{strings,refs}

\end{document}